\documentclass[sigconf,natbib=false]{acmart}
\AtBeginDocument{%
  }

\setcopyright{acmlicensed}
\copyrightyear{2026}
\acmYear{2026}
\acmDOI{XXXXXXX.XXXXXXX}
\acmConference[CAIS '26]{ACM Conference on AI and Agentic Systems}{May 26--29,
  2026}{San Jose, CA}
\acmISBN{978-1-4503-XXXX-X/2018/06}

\RequirePackage[
  datamodel=acmdatamodel,
  style=acmnumeric,
  ]{biblatex}
\usepackage{xspace}
\usepackage{subcaption}
\usepackage{enumitem}

\newenvironment{myitemize}{%
\begin{itemize}[leftmargin=1em, itemsep=.1em, parsep=.1em, topsep=.1em,
    partopsep=.1em]}
{\end{itemize}}

\newcounter{usecase}
\newcommand{\UseCase}[2]{%
  \refstepcounter{usecase}%
  \noindent\textbf{Use Case~\theusecase: #1}\label{#2}
}

\newcounter{challenge}
\newcommand{\Challenge}[2]{%
  \refstepcounter{challenge}%
  \noindent\textbf{Challenge~\thechallenge: #1}\label{#2}
}

\newcommand{\ps}{\textsc{Pneuma-Seeker}\xspace}
\newcommand{\conductor}{\textsc{Conductor}\xspace}
\newcommand{\materializer}{\textsc{Materializer}\xspace}
\newcommand{\retriever}{\textsc{Retriever}\xspace}
\newcommand{\pr}{\textsc{Pneuma-Retriever}\xspace}
\newcommand{\smolagent}{\textsc{smolagents}\xspace}
\newcommand{\dsguru}{\textsc{DS-Guru}\xspace}

\newcommand{\pneuma}{\textsc{Pneuma}\xspace}
\newcommand{\octopus}{\textsc{Octopus}\xspace}

\newcommand{\duckdb}{\textsc{DuckDB}\xspace}
\newcommand{\dbservice}{\textsc{DBService}\xspace}
\newcommand{\lmservice}{\textsc{LMService}\xspace}
\newcommand{\ts}{$(\mathcal{T},S)$\xspace}

\newcommand{\latentInfoNeed}{$\mathcal{I}^*$\xspace}
\newcommand{\activeInfoNeed}{$\mathcal{I}^+$\xspace}
\newcommand{\outputDocument}{$\mathcal{D}$\xspace}

\newcommand{\tinyskip}{\vspace{3pt}}
\newcommand{\mypar}[1]{\tinyskip\noindent\textbf{#1.}\xspace}

\begin{document}

\title{Demonstration of Pneuma-Seeker: Agentic System for Reifying and Fulfilling Information Needs on Tabular Data}
\newcommand{\uchicago}{
  \institution{The University of Chicago}
  \city{Chicago}
  \country{USA}
}

\author{Muhammad Imam Luthfi Balaka}
\email{luthfibalaka@uchicago.edu}
\orcid{0009-0001-5324-7758}
\affiliation{\uchicago}

\author{Raul Castro Fernandez}
\email{raulcf@uchicago.edu}
\orcid{0000-0001-7675-6080}
\affiliation{\uchicago}

\begin{abstract}
Data analysts working with relational data often start with vague or underspecified questions and refine them iteratively as they explore the data. To support this iterative process, we demonstrate \ps, a system that reifies a user's information need as explicit, inspectable relational specifications, enabling iterative refinement of the information need, targeted data discovery, and provenance-aware execution. Through two real-world procurement use cases, we show how \ps leverages LLMs as transparent, interactive analytical collaborators rather than opaque answer engines.
\end{abstract}



\received{13 March 2026}
\received[revised]{13 March 2026}
\received[accepted]{13 March 2026}

\maketitle

\section{Introduction}
\label{sec:intro}
Organizations increasingly rely on data to inform operational, financial, and policy decisions. A typical workflow begins with a stakeholder articulating an information need. Data workers (e.g., analysts and domain experts) then identify and prepare the data needed to address it. In practice, the workflow is rarely linear. Stakeholders refine requests as new information emerges, and data workers revise the corresponding data discovery and preparation steps. We identify several key challenges in this process, informed by our collaboration with the procurement office at our university.

\begin{myitemize}
    \item \Challenge{Answer Correctness}{ch:answer_correctness}
    The resulting answer must be correct to support reliable decision making. Correctness depends on multiple factors, including identifying relevant data sources, performing correct data preparation (e.g., joins and cleaning), and respecting domain constraints or assumptions governing the interpretation of the data.

    \item \Challenge{Imprecise Information Needs}{ch:vague_info_need}  
    Obtaining the correct answer requires a precise specification of the information need. However, users rarely articulate this specification completely at the outset. Relevant variables, filters, or constraints may only become apparent after inspecting the data. We therefore distinguish between two notions of information need. The \textit{latent information need} \latentInfoNeed denotes the fully specified characterization of the data required to solve a task. In contrast, an \textit{active information need} \activeInfoNeed represents the user's current articulation of \latentInfoNeed at a particular point in time.

    \item \Challenge{Inspectability}{ch:provenance} 
    Because \activeInfoNeed may evolve over time, inspection and verification of intermediate results are essential. Users must be able to examine how candidate data was discovered and integrated. This transparency helps identify errors, resolve ambiguities, and ensure that the final result fulfills \activeInfoNeed.

    \item \Challenge{Latency}{uc:time}
    Collaborative data work is communication-intensive. Stakeholders and data workers repeatedly clarify intent, evaluate intermediate results, and revise analyses. Consequently, resolving a single task may take weeks or months.
\end{myitemize}

Recent advances in large language models (LLMs) demonstrate strong capabilities in interpreting natural-language instructions and performing structured reasoning~\cite{CoT2022,LLMZeroShotReasoners2022,yao2023reactsynergizingreasoningacting}. These capabilities have motivated systems that leverage LLMs to automate analytical workflows over tabular data. For example, \dsguru~\cite{lai2025kramabenchbenchmarkaisystems} prompts an LLM to decompose a question into subtasks, reason through intermediate steps, and synthesize executable Python code implementing the derived plan. Agentic frameworks such as \smolagent~\cite{smolagents2024} provide tool-calling interfaces that allow an LLM to inspect table schemas and content, execute code, and iteratively refine its reasoning through interaction with data. Other recent systems explore related uses of LLMs for data processing. For example, \textsc{Chain-of-Table}~\cite{wang2024chainoftable} studies stepwise reasoning over a single table, while \textsc{Palimpzest}~\cite{LiuPalimpzest2025} and \textsc{DocETL}~\cite{DocETL2025} investigate LLM-driven pipelines for transforming semi-structured or unstructured data. These efforts highlight the growing interest in applying LLMs to automate data-centric workflows.

These approaches demonstrate that LLMs can perform multi-step analytical procedures. However, they share a common structural assumption: the user's question is already sufficiently precise. Under this assumption, the question is typically interpreted implicitly by the model, translated into reasoning traces or executable code, and executed directly. The operational specification of the task remains embedded within generated code or intermediate reasoning rather than represented as an explicit artifact that users can inspect independently of execution. Consequently, semantic misalignment between the user's intended meaning and the model's interpretation may only become apparent after execution.

This observation motivates the key idea of \ps~\cite{balaka2026pneumaseekerrelationalreificationmechanism}: the interpretation of user intent should be externalized into a concrete, inspectable artifact separate from execution. We therefore present \ps, an agentic system for interactive data discovery and preparation that reifies a user's \activeInfoNeed as an explicit relational schema. Rather than treating a question as a prompt whose output is an answer, \ps treats it as a specification that must be progressively refined. The system proposes a set of target relations that sufficiently address \activeInfoNeed.

The resulting set of relations functions as a shared contract between user and system. Given this set of relations, the system identifies relevant tables and materializes the specified relations. Through iterative refinement of the relations, users can progressively converge from \activeInfoNeed to \latentInfoNeed.

We demonstrate two representative use cases derived from discussions with our university's procurement office:

\begin{myitemize}
    \item \UseCase{Outstanding Amount}{uc:outstanding}
   \textit{ ``I want to understand how much money is still outstanding for our purchase orders. Can you compute the total open amount for FY2025?'' } 
    This use case highlights ambiguity in financial terminology and illustrates how the system supports the user in refining the semantics of the term ``outstanding.'' The task uses real procurement data extracted from JAGGAER,\footnote{\url{https://www.jaggaer.com}} a widely used procurement platform in U.S. higher education. The dataset contains 41 tables with a total size of 15~GB. We additionally incorporate a FY2025 purchase order table extracted from an \textsc{Oracle} database (7~MB).

    \item \UseCase{Test Outsourcing}{uc:outsourcing}
    \textit{``Does outsourcing tests to external vendors potentially yield cost savings?''}
    In this scenario, the user brings their own tables to evaluate whether laboratory tests should remain in-house or be outsourced. Two tables are provided: an internal test table with schema \textit{[EAP, CPT, EAP Description, FY25 Price, FY26 Price]} and a vendor proposal table with schema \textit{[Vendor\_1 Bill Code, Vendor\_1 Test Description, Vendor\_1 Price, Vendor\_2 Bill Code, ...]}. The tables do not share common identifiers, and test descriptions vary in granularity and terminology. As a result, equality joins are not applicable and naive string matching fails to recover meaningful correspondences. This use case demonstrates how \ps supports semantic operations, in particular semantic join.
\end{myitemize}

Together, these use cases demonstrate how \ps converts underspecified \activeInfoNeed into explicit relational specifications that support grounded execution, user validation, and iterative refinement, enabling users to converge toward \latentInfoNeed.

\section{System Overview}
\label{sec:sys_overview}

\ps reifies \activeInfoNeed as $(\mathcal{T}, S)$, where $\mathcal{T}$ is a set of views derived from an underlying table collection $\mathcal{C}$, and $S$ is a Python script over $\mathcal{T}$ that produces the output document \outputDocument to fulfill \activeInfoNeed.

\begin{figure}[h]
    \centering
    \includegraphics[width=\linewidth]{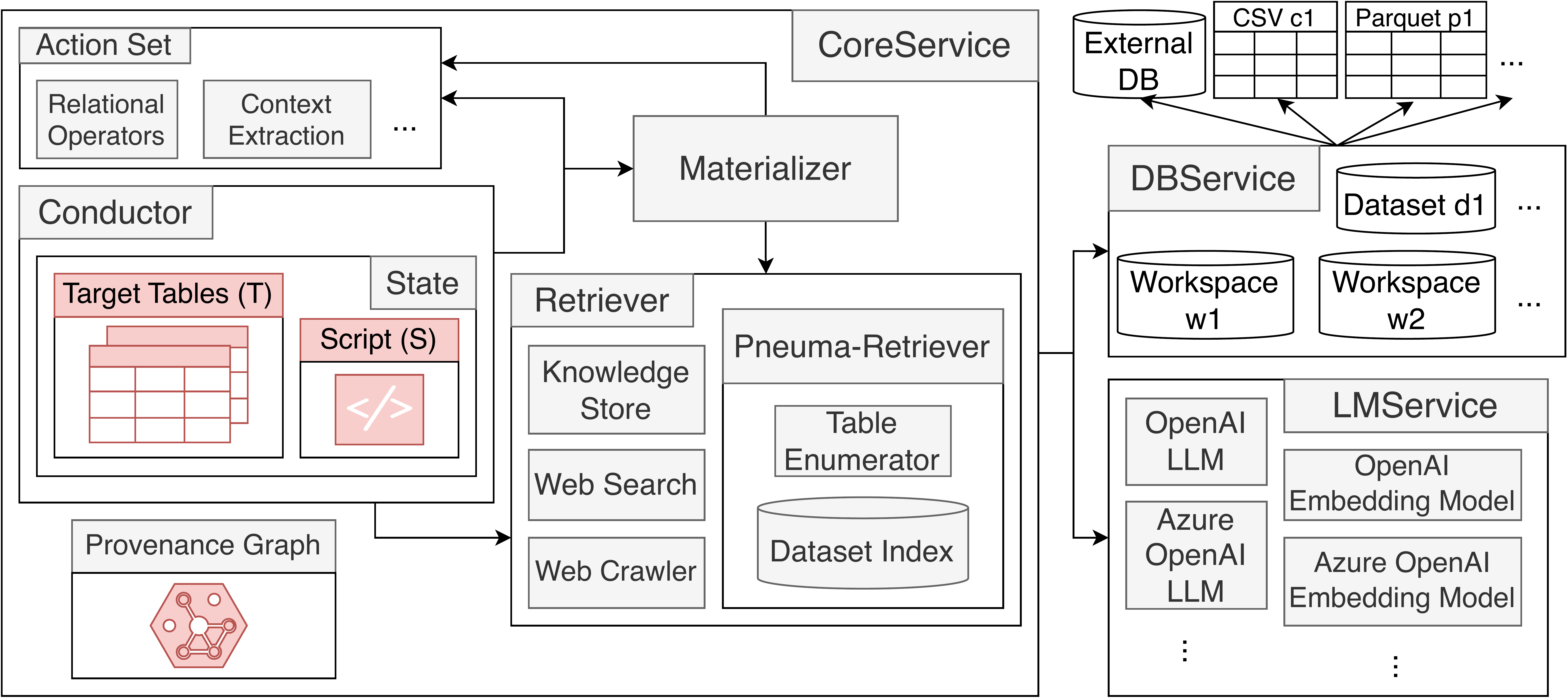}
    \caption{Architecture of \ps}
    \label{fig:architecture}
\end{figure}

Figure~\ref{fig:architecture} illustrates the architecture of \ps, which consists of three core components: \conductor, \materializer, and \retriever. \conductor orchestrates the workflow. It translates \activeInfoNeed into $(\mathcal{T}, S)$, plans actions, invokes retrieval and materialization, and manages user interaction. \materializer constructs the views in $\mathcal{T}$ by applying relational and semantic operators, producing intermediate tables tracked in a provenance graph. \retriever primarily performs data discovery over $\mathcal{C}$ and returns relevant tables.

The core components are supported by two services: \dbservice and \lmservice. \dbservice uses \duckdb~\cite{RaasveldtDuckDB2019} to enable persistence of \ts and query execution. Each user session operates within an isolated workspace, where $\mathcal{C}$ is dynamically attached to the workspace. \lmservice provides a unified interface to LLMs and embedding models.

\mypar{Context Management}
The decomposition of \ps into multiple components reflects a macro-level context management strategy. Each component operates within a scoped responsibility (orchestration, retrieval, or materialization), so that no single component must reason over the entire context of the workflow. This modularization reduces prompt size and mitigates performance degradation observed with long contexts~\cite{context-length-hurts-llm}, while also constraining the action spaces of the components.

However, even after retrieval, directly placing full tables into the LLM context is infeasible. For example, a purchase order table in the dataset of Use Case~\ref{uc:outstanding} with 117 columns and 5 million rows would require at least 2.1 billion tokens, far exceeding the 200{,}000-token limit of the model we use (\verb|o3-2025-04-16|). Conversely, naive sampling risks omitting critical evidence.

To address this, \ps introduces a complementary \emph{micro-level} context management mechanism. In addition to receiving row samples, the model may issue executable queries through \dbservice to probe retrieved tables, e.g., verifying value existence, computing distributions, or inspecting structural properties. This protocol enables targeted evidence extraction while maintaining scalability and directly supports correctness by allowing the system to resolve structural irregularities that sampling-based approaches may miss. Although the model could in principle perform such checks, explicitly modeling this capability as an action encourages it to, e.g., verify assumptions before committing to interpretations.

\mypar{Dynamic Planning}
Both \conductor and \materializer operate within bounded planning loops around model inference. Each iteration begins with \textbf{Situational Analysis}, which evaluates the current state (e.g., \ts) and prior actions to select a sequence of actions for the iteration. Planning terminates when either a user-facing response is produced or a predefined limit is reached.

\mypar{Materialization}
Given a specification of $\mathcal{T}$, \materializer constructs it using the following actions:

\begin{myitemize}
    \item \textbf{Relational Operators:} dedicated join, union, and projection operations. These operators are less prone to errors than LLM-generated queries or Python code.
    \item \textbf{Semantic Operators:} we currently implement the following:
    \begin{myitemize}
        \item Semantic Join: tuple matching using embedding and syntactic similarity between textual attributes, inspired by prior embedding-based dataset linking approaches~\cite{SeepingSemantics2018} and semantic operator frameworks such as LOTUS~\cite{PatelSemanticOperator2024}.
        \item Semantic Column Generation: synthesizing new attributes conditioned on existing columns, inspired by \textsc{Palimpzest}~\cite{LiuPalimpzest2025}.
    \end{myitemize}
    \item \textbf{Query Executor:} custom SQL when operators are insufficient.
    \item \textbf{Python Executor:} Python code with access to \dbservice APIs as fallback for maximally expressive transformations.
\end{myitemize}

All transformations are recorded in a provenance DAG. A topological ordering of this graph yields a deterministic script that reconstructs $\mathcal{T}$ and executes $S$ to produce \outputDocument.

\mypar{Retriever}
\retriever provides a unified interface for retrieval mechanisms. The primary retriever, \pr, combines schema-level table discovery via \pneuma~\cite{BalakaPneuma2025} with content-aware keyword matching inspired by \octopus~\cite{LiOctopus2026}. Additional retrievers include a Knowledge Store for accumulated domain knowledge~\cite{park-etal-2021-scalable,liu-etal-2025-user,CaptureKnowledgeUrban2025}, as well as Web Search and Web Crawl.

\section{Demonstration}
\label{sec:demo}

\begin{figure*}[t]
    \centering
    \begin{subfigure}{0.49\linewidth}
        \centering
        \includegraphics[width=\linewidth]{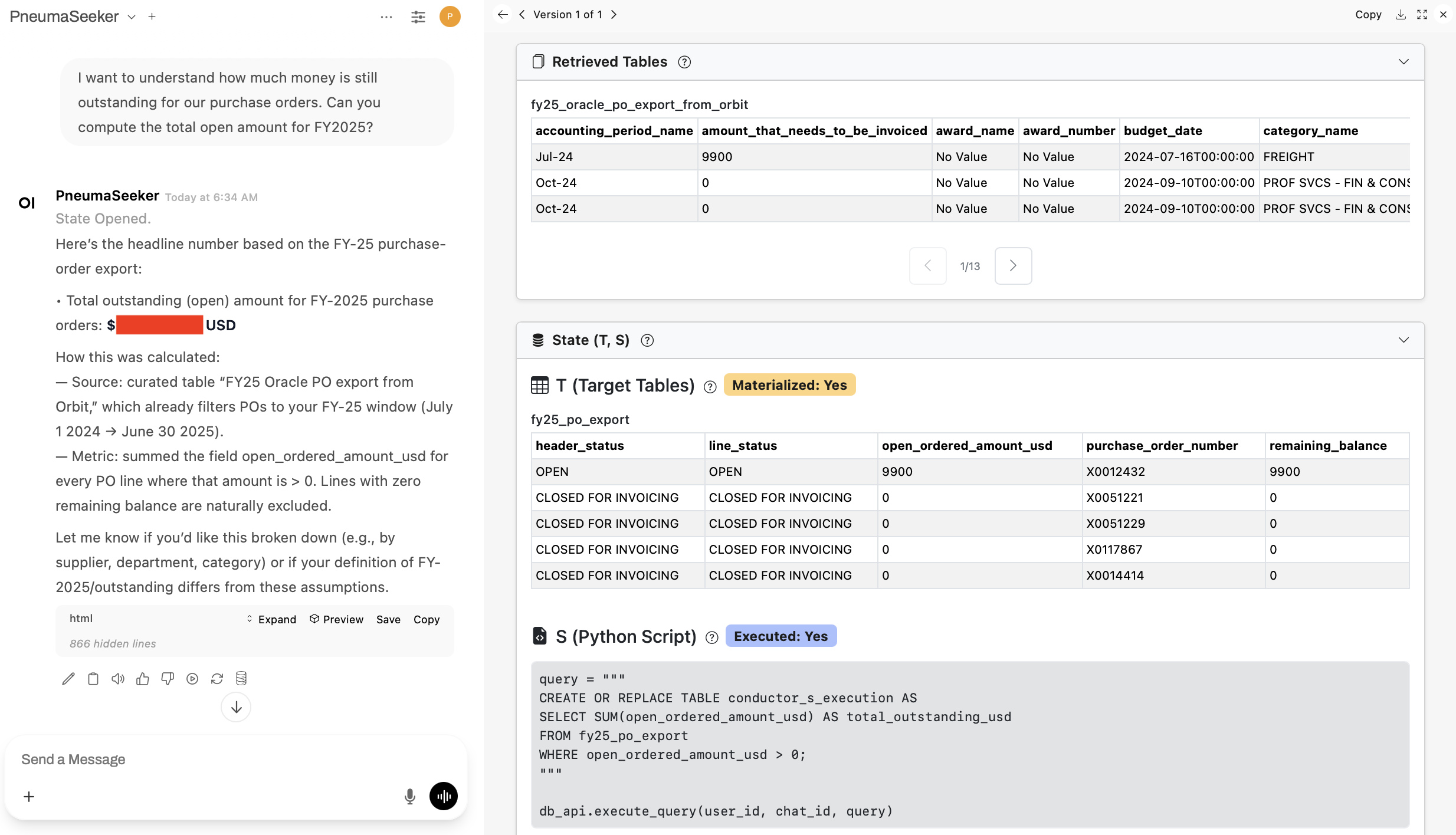}
        \caption{\ps on Use Case~\ref{uc:outstanding}: Outstanding Amount}
        \label{fig:demo_outstanding}
    \end{subfigure}
    \hfill
    \begin{subfigure}{0.49\linewidth}
        \centering
        \includegraphics[width=\linewidth]{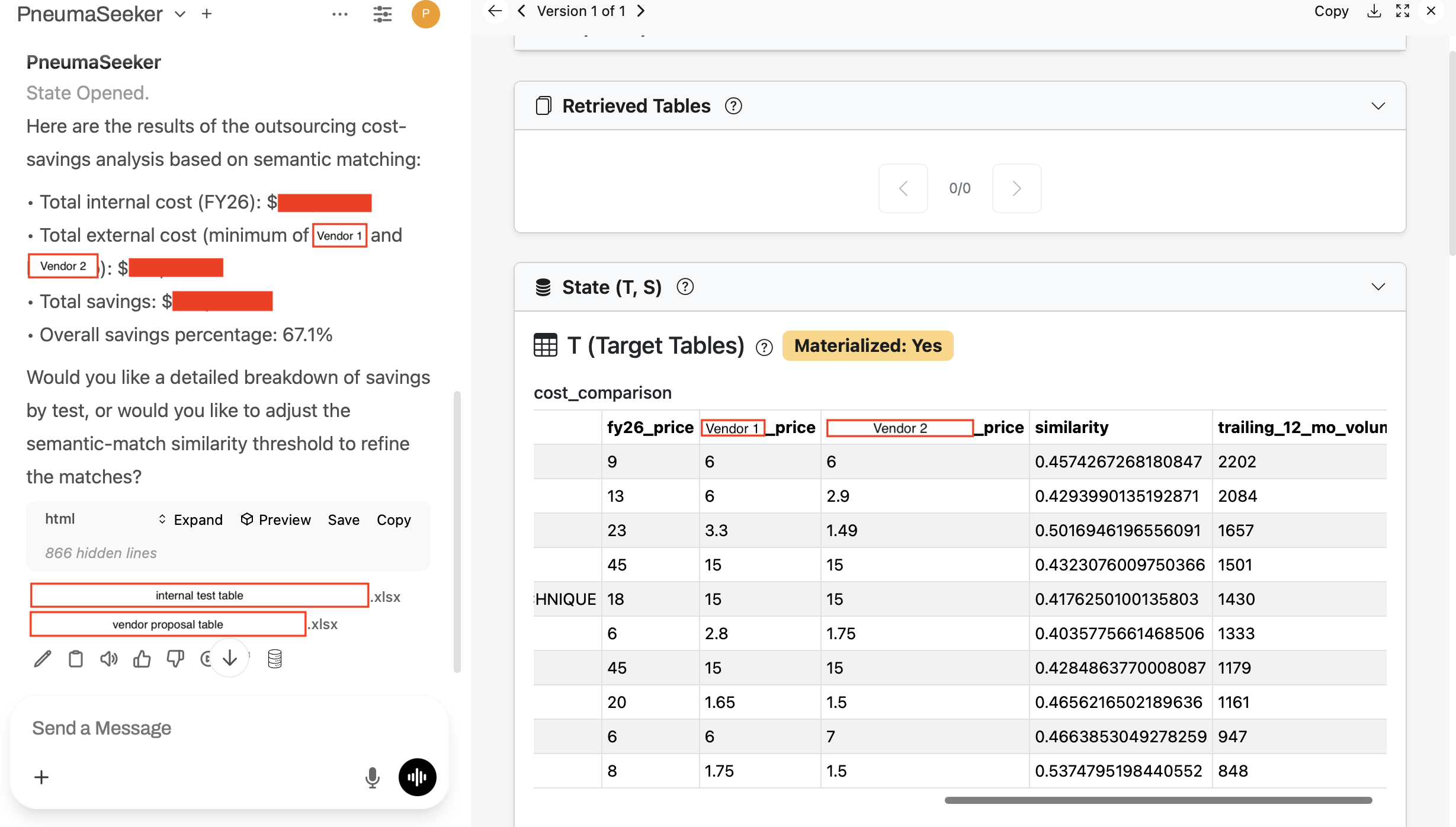}
        \caption{\ps on Use Case~\ref{uc:outsourcing}: Test Outsourcing}
        \label{fig:demo_outsourcing}
    \end{subfigure}

    \caption{Demonstration of \ps on Two Use Cases}
    \label{fig:demo}
\end{figure*}

We demonstrate \ps through the two procurement use cases introduced in Section~\ref{sec:intro}. The demonstration illustrates how relational reification supports inspectability, iterative refinement, and grounded execution over a collection of tables. During the session, participants will interact with \ps through its interface, which consists of a chat-based interaction panel and a sidebar that exposes elements such as \ts and provenance of $\mathcal{T}$.

Because the procurement data used in our use cases contains sensitive institutional information, the live demonstration will instead use datasets derived from Chicago Open Data~\cite{chicago_data_portal_2026}. Participants will be able to explore these datasets through \ps using a set of example \activeInfoNeed prompts. The interface allows participants to inspect \ts and iteratively refine their requests.

\subsection{Use Case~\ref{uc:outstanding}: Outstanding Amount}

This use case illustrates how \ps supports schema inspection and semantic refinement when computing financial aggregates. The system produces an answer in 2 minutes and 24 seconds (Figure~\ref{fig:demo_outstanding}).

\mypar{Step 1: Retrieve Relevant Tables}
The interaction begins when the user poses the question: \textit{``What is the total outstanding amount across open purchase orders for FY2025?''} At this stage, no tables have yet been retrieved for the session. \ps therefore invokes \pr to identify candidate tables relevant to the \activeInfoNeed. Among the retrieved tables are \textit{fy25\_oracle\_po\_export\_from\_} \textit{orbit} and \textit{purchase\_order\_line}, both of which contain purchase order information.

To determine which table best corresponds to the requested time period, \ps issues a query to examine the date ranges present in both datasets. The table \textit{fy25\_oracle\_po\_export\_from\_orbit} contains records primarily from 2024 onward and is explicitly labeled with the identifier ``FY25,'' suggesting that it corresponds to fiscal year 2025 data. In contrast, the table \textit{purchase\_order\_line} contains older records covering earlier time periods. Based on these signals, \ps considers \textit{fy25\_oracle\_po\_export\_from\_orbit} as the reference table to form $\mathcal{T}$.

\mypar{Step 2: Define \ts, Materialize $\mathcal{T}$, Execute $S$}
Next, \ps reifies the \activeInfoNeed into \ts. In this case, $\mathcal{T}$ consists of a single derived view, \textit{fy25\_po\_export}, whose schema is: \textit{[header\_status, line\_status, open\_ordered\_amount\_usd, purchase\_order\_number, remaining\_balance]}.

The view is materialized as a projection over \textit{fy25\_oracle\_po\_export\_} \textit{from\_orbit}, exposing attributes relevant to computing outstanding purchase order amounts. The script $S$ performs an aggregation over the attribute \textit{open\_ordered\_amount\_usd} to compute the total outstanding amount.

After materializing $\mathcal{T}$, \ps executes $S$ and produces a natural-language answer summarizing the computed total.

\mypar{Step 3: Inspection and Refinement of \ts}
Although the system produces a numerical result, the user may wish to verify the assumptions underlying the computation. The sidebar interface enables inspection of retrieved tables and \ts.

During inspection, the user may notice that \ps implicitly interprets the notion of ``outstanding'' as the attribute \textit{open\_ordered\_amount\_usd}. However, the derived view also contains another attribute, \textit{remaining\_balance}, which could plausibly represent the same concept. In addition, the underlying source table includes a further attribute, \textit{amount\_that\_needs\_to\_be\_invoiced}, which was not projected into the derived view. Inspection of sample values reveals that these attributes exhibit similar magnitudes, suggesting that they may correspond to related financial quantities.

These observations prompt the user to reconsider the precise definition of ``outstanding'' in this context. Rather than relying solely on the model's initial interpretation, the user can refine the \activeInfoNeed by specifying which attribute should represent the intended financial measure. Importantly, such ambiguities would be difficult to identify from the chat interface alone, but become apparent through inspection of the schema and sample values exposed in the sidebar.

A second potential source of ambiguity concerns the interpretation of ``FY2025.'' The system assumes this definition: July 1, 2024 through June 30, 2025. However, the user may wish to verify whether this convention matches the convention in their office. Furthermore, because $S$ simply aggregates over the derived view, the user may want to confirm that the underlying dataset indeed contains only records within the intended date range. These questions cue the user to further refine \activeInfoNeed.

\subsection{Use Case~\ref{uc:outsourcing}: Test Outsourcing}

This use case illustrates how \ps supports semantic join when tables do not share common identifiers. The system produces an initial response in 1 minute and 41 seconds. After refinement of the integration strategy, a second response is produced in 3 minutes and 27 seconds (Figure~\ref{fig:demo_outsourcing}).

For clarity, we refer to the two input tables as the \textit{internal test table} and the \textit{vendor proposal table}. The internal table contains laboratory tests performed in-house, including attributes such as \textit{EAP}, \textit{CPT}, and internal pricing. The vendor proposal table lists externally provided tests and associated prices from multiple vendors.

\mypar{Step 1: Retrieve Relevant Tables}
The user poses the question: \textit{``Does outsourcing tests to external vendors potentially yield cost savings?''} Unlike the previous use case, the user uploads the relevant spreadsheets at the beginning of the interaction. Because the uploaded tables already correspond directly to the \activeInfoNeed, \ps decides to not invoke \pr. Instead, it treats the uploaded tables as the primary sources to form $\mathcal{T}$.

\mypar{Step 2: Define \ts, Materialize $\mathcal{T}$, Execute $S$}
\ps first attempts to construct a derived view named \textit{cost\_comparison}. The view is defined by joining the internal test table with the vendor proposal table using test identifiers. The goal of this join is to align each internal test with the corresponding entries from the vendor proposals, enabling direct comparison of prices.

However, the identifiers used in the two tables follow different conventions. For example, internal tests use CPT codes such as ``86317,'' while the vendor proposal table contains codes such as ``0000718.'' Because these identifiers do not share the same semantics or formatting, the equality join produces an empty result. Consequently, the system outputs the following:

\textit{The analysis returned zero matched tests and thus zero calculated savings. It appears our CPT-based join did not find any overlaps between internal codes and the external bid codes (e.g., ``86317'' vs.\ ``0000718''). To accurately assess outsourcing savings, we may need to realign the code formats (for example, zero-padding internal CPTs) or perform a semantic join on test descriptions.}

The user then requests a semantic join based on test descriptions. In response, \ps constructs an alignment between tests using textual similarity and semantic matching. This refined integration strategy produces a non-empty join result and enables the system to compute potential cost differences between internal and external pricing.

\mypar{Step 3: Inspection and Refinement of \ts}
The sidebar interface allows the user to inspect the derived \textit{cost\_comparison} view. Each row contains the internal test price (e.g., \textit{fy26\_price}) alongside the corresponding vendor prices (\textit{vendor1\_price} and \textit{vendor2\_price}). 

Inspection reveals that some matches appear reasonable. For example, a test priced at \$9 internally may correspond to vendor prices of \$6, suggesting potential savings through outsourcing. However, other matches exhibit large discrepancies, such as an internal price of \$23 paired with vendor prices of \$3.30 and \$1.49. These discrepancies raise questions about whether the matched tests are truly equivalent or whether differences in naming conventions or test granularity have produced incorrect alignments.

This example highlights an important limitation of semantic matching: while it can recover useful correspondences when explicit identifiers are absent, it may also introduce imperfect matches. Nevertheless, the derived view provides a concrete starting point for further analysis. By inspecting the candidate alignments, users can identify questionable matches, refine the integration strategy, or incorporate additional constraints. In this way, \ps enables users to iteratively refine \activeInfoNeed while maintaining visibility into the intermediate results that shape the final result.

\section{Conclusion}
\label{sec:conclusion}
We demonstrated \ps, a system that reifies a \activeInfoNeed as explicit relational specifications \ts to support interactive data discovery and preparation, on two real-world procurement use cases. These examples illustrate how \ps allows users to interact an with LLM-powered system as a transparent analytical collaborator while maintaining visibility into the data transformations that produce the final results, helping them converge toward~\latentInfoNeed.

\printbibliography

\end{document}